%% file: main.tex
\documentclass[11pt]{article}

\usepackage[preprint]{acl}

\usepackage{times}
\usepackage{latexsym}

\usepackage[T1]{fontenc}

\usepackage[utf8]{inputenc}

\usepackage{microtype}

\usepackage{inconsolata}

\usepackage{graphicx}

\usepackage{microtype}
\usepackage{hyperref}
\usepackage{url}
\usepackage{booktabs}

\usepackage{lineno}

\usepackage{tabularx}
\usepackage{graphicx}
\usepackage{enumitem}
\usepackage{xspace}
\usepackage{subcaption}
\usepackage{siunitx}
\usepackage[table]{xcolor}
\usepackage{fancyvrb}   %
\usepackage{fvextra}
\usepackage{multirow}
\usepackage{multicol}

\usepackage{amsmath}
\usepackage{xcolor}
\definecolor{mydarkblue}{rgb}{0,0.08,0.45}
\usepackage{wasysym}
\usepackage{hyperref}

\usepackage{algorithm}
\usepackage{algpseudocode}

\usepackage{tcolorbox}
\tcbuselibrary{breakable,skins}

\usepackage{xcolor}

\definecolor{darkblue}{rgb}{0, 0, 0.5}
\hypersetup{colorlinks=true, citecolor=darkblue, linkcolor=darkblue, urlcolor=darkblue}

\newtcolorbox{LLMPrompt}[1][]{
  enhanced, breakable, verbatim,
  colback=gray!5, colframe=black!15, boxrule=0.4pt, arc=1.5mm,
  left=6pt,right=6pt,top=6pt,bottom=6pt,
  fontupper=\small\ttfamily,
  colbacktitle=gray!50,
  #1
}
\newtcolorbox{LLMPromptSmall}[2][]{
  enhanced, verbatim,
  colback=gray!5, colframe=black!15, boxrule=0.4pt, arc=1.5mm,
  left=6pt,right=6pt,top=6pt,bottom=6pt,
  fonttitle=\fontsize{9}{11}\normalfont,
  fontupper=\fontsize{7}{9}\selectfont\ttfamily,
  colbacktitle={#2!50},
  before upper=\setlength{\parskip}{0pt}\setlength{\parindent}{0pt}\ignorespaces,
  after  upper=\unskip\par,
  #1
}

\newcommand{\bench}{\texttt{CRAB-Bench}\xspace}
\newcommand{\user}{\texttt{RUSE}\xspace}

\title{\raisebox{-0.3\height}{\includegraphics[height=2.5em]{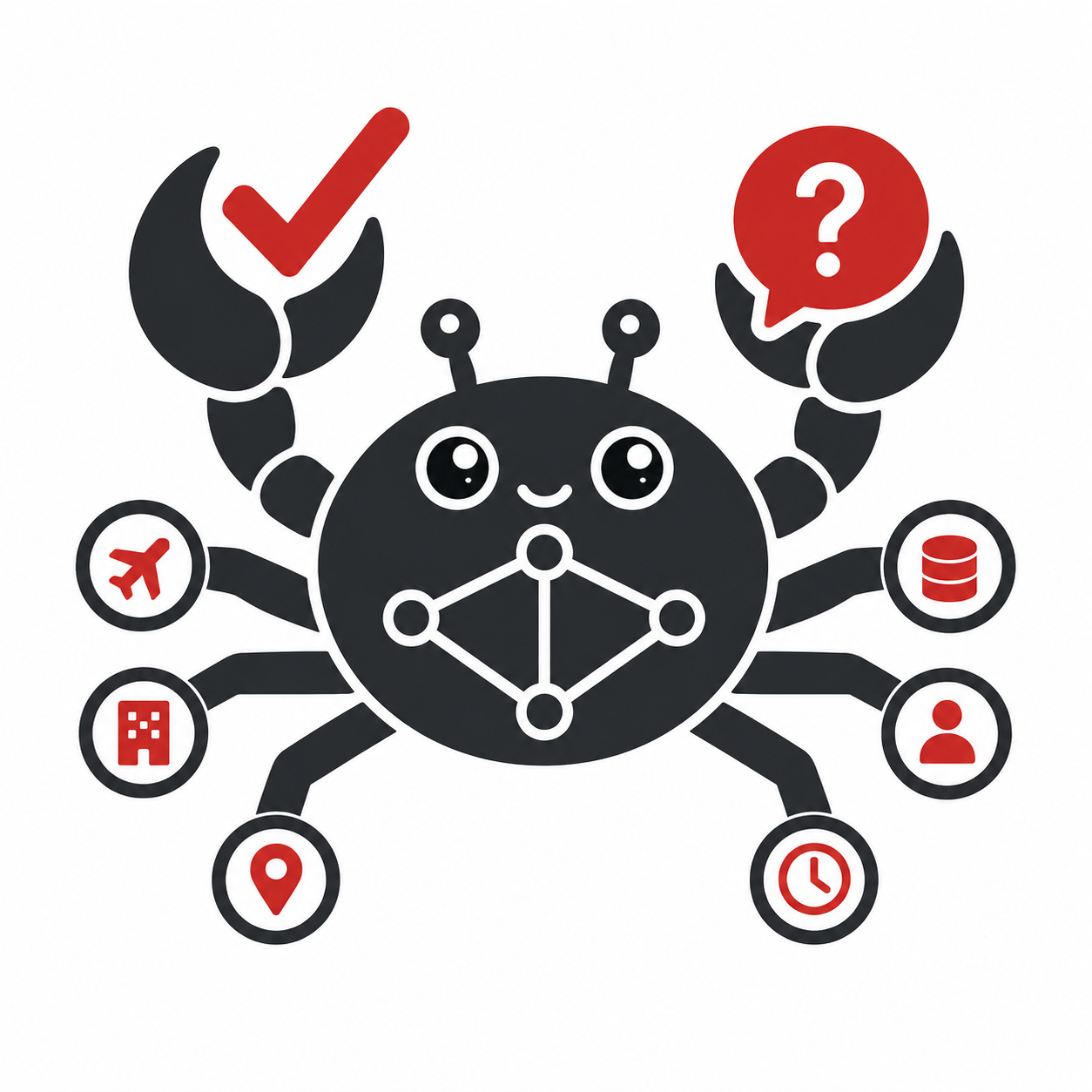}}~CRAB-Bench: Evaluating LLM Agents under
  Complex Task Dependencies and Human-aligned User Simulation}

\author{
    \textbf{Danqing Wang\textsuperscript{1}},
    \textbf{Akshay Sivaraman\textsuperscript{1}},
    \textbf{Lei Li\textsuperscript{1}},
\\
     \textsuperscript{1}Carnegie Mellon University,
\\
 \small{
   \href{mailto:danqingw@cs.cmu.edu}{danqingw@cs.cmu.edu}
 }
}

\begin{document}
\maketitle

\begin{abstract}
\input{sec/000abstract}
\end{abstract}

\section{Introduction}
\label{sec:introduction}
\input{sec/001introduction}

\section{Related Work}
\label{sec:related}
\input{sec/002related}

\section{Constraint-based Realistic Agent Benchmark}
\label{sec:benchmark}
\input{sec/003benchmark}

\section{Experimental Results}
\label{sec:experiment}
\input{sec/004evaluation}

\section{Analysis}
\label{sec:analysis}
\input{sec/005analysis}

\section{Conclusion}
\label{sec:conclusion}
\input{sec/006conclusion}

\section*{Limitations}
\label{sec:limitation}
\input{sec/020limitation}

\bibliography{reference}

\newpage
\appendix

\section{Appendix}
\label{sec:appendix}
\input{sec/010appendix}

\end{document}

%% file: sec/000abstract.tex
Evaluating LLM agents in realistic service scenarios requires complex task dependencies, imperfect user behavior, and an evaluation that accommodates multiple valid solutions.
We introduce \bench (Constraint-based Realistic Agent Benchmark) and \user (Realistic User Simulation Engine) to address this gap.
\bench generates tasks via a constraint graph over multiple interdependent entities with structured distractors, requiring agents to reason carefully over thousands of misleading candidates where only a tiny fraction of solutions are valid.
\user replaces cooperative, template-like simulators with realistic users grounded in human behavioral studies, instantiated across diverse personas and four behavioral dimensions.
Experiments on four frontier LLM agents show that the best model achieves only 61\% pass@1 on \bench, and switching to \user causes further drops of up to 57\%, concentrated in task-solving ability rather than conversational quality.
Information Disclosure is the most damaging behavioral dimension, and agents interacting with \user are less likely to admit mistakes, instead masking errors through implicit corrections.

%% file: sec/001introduction.tex
LLM-based agents are increasingly deployed in complex, multi-turn service scenarios such as travel planning, customer support, and technical assistance~\citep{chiang2024chatbot,zhang2025survey,yao2025taubench,barres2025tau2}. Systematically evaluating agents in these settings is challenging for three reasons.
First, real user requests involve \emph{multi-step task dependencies}: satisfying one subtask implicitly constrains others, requiring the agent to plan across entities and propagate consequences without being explicitly told to do so.
Second, real users are \emph{imperfect}: they disclose information incrementally, communicate ambiguously, and react emotionally to errors~\citep{laban2025llms,qian2025userbench,zhou2026mind}.
Third, most tasks admit \emph{multiple valid solutions}, so evaluation cannot simply compare agent output against a single ground-truth record.

\begin{figure}[t]
    \centering
    \includegraphics[width=1\linewidth]{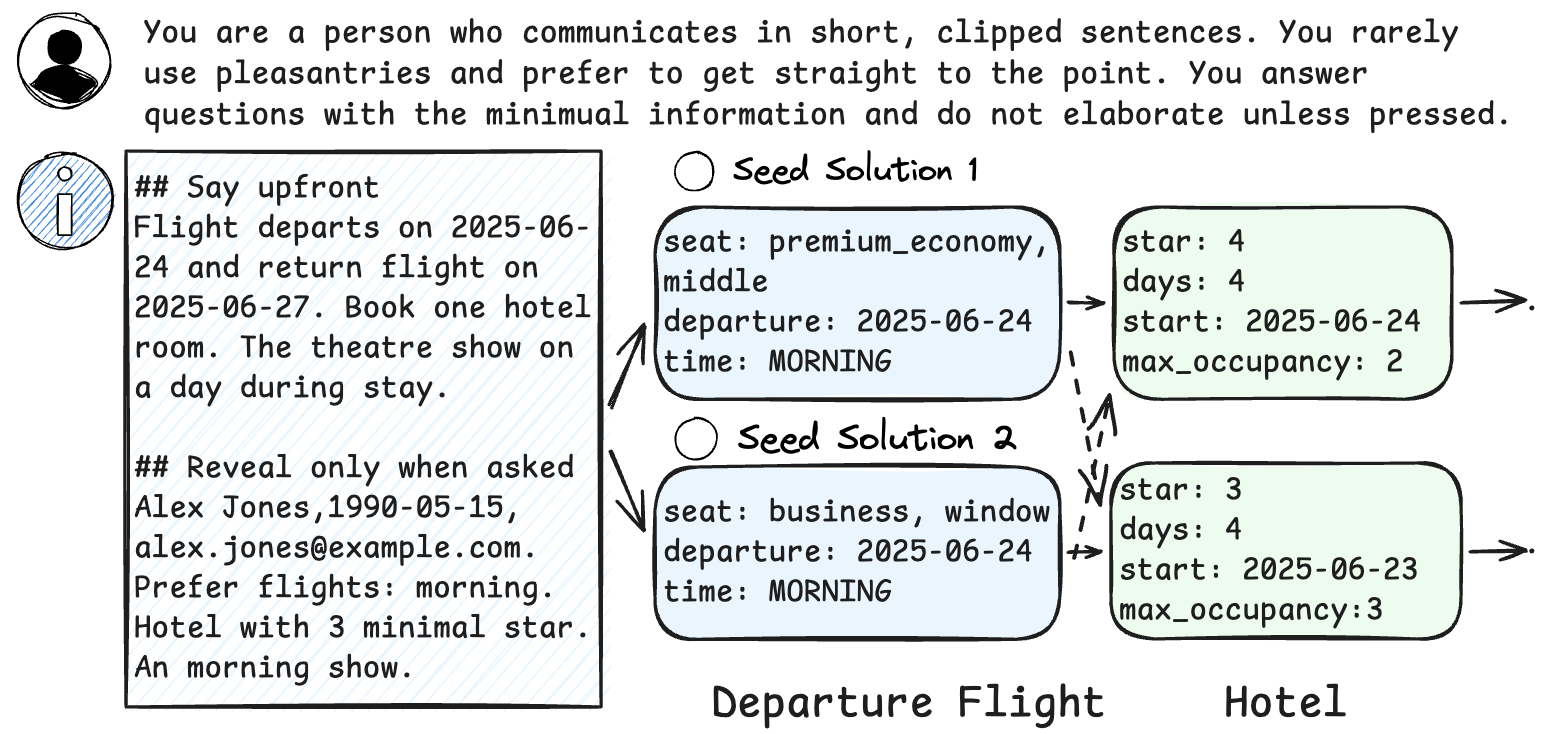}
    \caption{One task in \bench with user persona and information control. 16 solutions satisfy the user requirements due to the combinations of different parts of seed solutions (as shown in the dotted lines). }
    \label{fig:task}
\end{figure}

Existing benchmarks address these challenges only partially.
Single-tool and single-interaction benchmarks~\citep{qintoolllm,guo2024stabletoolbench,liuagentbench,jimenezswe,vero2025baxbench} omit inter-task dependencies and multi-turn dynamics.
Multi-turn benchmarks such as $\tau$-bench~\citep{yao2025taubench} and $\tau^2$-bench~\citep{barres2025tau2} use cooperative, template-like user simulators and rigid ground-truth evaluation, and have already been largely saturated.
Meanwhile, recent studies show that LLM-based user simulators systematically miscalibrate evaluation and diverge substantially from real human behavior~\citep{seshadri2026lost,zhou2026mind}, raising doubts about the gaps between benchmark performance and realistic user scenarios.

To address these gaps, we propose \bench (Constraint-based Realistic Agent Benchmark), an agentic evaluation framework with 3 components.
\begin{itemize}[topsep=0pt,itemsep=2pt,parsep=0pt,leftmargin=1.5em]
    \item \textbf{Constraint graph-based task generation.}
    We model each task as a constraint graph over subtask nodes with domain constraints and edge constraints to capture real-world coherence requirements (e.g., a hotel check-in must follow a flight's arrival). We propose a constraint graph-based generation pipeline to automatically generate seed solutions via CSP  (Constraint Satisfaction Problem) solver and populate the database with distractors, ensuring both solvability and complexity. 

    \item \textbf{Human-aligned user simulation.}
    We propose \user (Realistic User Simulation Engine) instantiates user simulators along four human behavior dimensions combined into three personas. 

    \item \textbf{State-based evaluation.}
    We evaluate agents through two complementary rule-based verifier sets: \emph{concrete-state} verifiers that inspect whether the final database satisfies all task requirements, and \emph{abstract-state} verifiers that check communication factuality and correct task termination. 
\end{itemize}

We instantiate \bench on a trip booking domain with 19 agent tools and 7 user tools, yielding 200 tasks stratified by difficulty (S1--S4). In the hardest stratum (S1), the distractor ratio based on the hard misleading distractors is only 0.05\%, and the full search space is more than $500^4$. 

We evaluate four LLM agents (\texttt{Claude Sonnet 4.6}, \texttt{DeepSeek V3.2}, \texttt{GLM-5}, \texttt{Qwen3 Coder Next}) and find: (i) increasing the number of distractors and task dependencies make the agents struggle with the misleading candidates; (ii) all models suffer consistent performance drops of 19--57\% when switching from a generic user simulator to \user; (iii) \user primarily degrades agents' ability to arrive at correct solutions (concrete-state drops of 17--39\%) rather than their conversational quality; (iv) Information Disclosure (D2) is the single most damaging behavioral dimension, and \texttt{Impatient} users, despite adversarial tone, provide useful error signals for problem-solving. We also provide a detailed failure analysis and suggest the possible improvement direction.

%% file: sec/002related.tex
\paragraph{Environments for User-centric Agents.}

Multi-turn agent benchmarks such as ToolSandbox~\citep{lu2025proactive}, URS~\citep{wang2024user}, MINT~\citep{wang2024mint}, $\tau$-bench~\citep{yao2025taubench}, and $\tau^2$-bench~\citep{barres2025tau2} evaluate agents in simulated service settings with tool use and dynamic user interaction. Recent $\tau-$-Banking~\citep{shi2026tau} focuses on unstructured knowledge in the financial domain, stress-testing the joint capability of retrieval and tool use in long-horizon, user-facing interactions. AgentChangeBench~\citep{rana2025agentchangebench} further studies robustness under mid-dialogue goal shifts.
However, these benchmarks share two limitations: they rely on cooperative, template-like user simulators, and they evaluate against fixed ground-truth solutions, making them unsuitable for tasks with multiple valid outcomes.
\bench addresses both by combining constraint graph-based task generation with structured distractors and state-based evaluation that explicitly accommodates open-ended solutions.

\paragraph{User Simulation for Conversational Agents.}
Early user simulation focuses on simulating user agents with different profiles or preferences, such as TravelPlanner+~\citep{singh-etal-2024-personal} and PRELUDE~\citep{gao2024aligning}. After that, more work focuses on the ambiguity in user interaction, such as Ambig-SWE~\citep{vijayvargiya2025interactive} and CharEval~\citep{li2026clareval}.
UserBench~\citep{qian2025userbench} introduces three core factors for realistic user simulators, who should share the information in an underspecified, incremental, and indirect way. 
However, recent studies also show that there are still significant gaps between the simulated users and the real-world human users, making the evaluation results less realistic. 
\citet{seshadri2026lost} find that evaluations using simulated users systematically miscalibrate: underestimating performance on challenging tasks and overestimating it on moderately difficult ones.
\citet{zhou2026mind} measures alignment between human and 31 LLM-based simulated users, finding that general LLM capability does not reliably translate to faithful user simulation.
Together, these results motivate the user simulation design choices of \bench: multiple user profiles with human-like interactive behaviors.

%% file: sec/003benchmark.tex
\begin{figure*}[thbp]
  \centering
  \includegraphics[width=\linewidth]{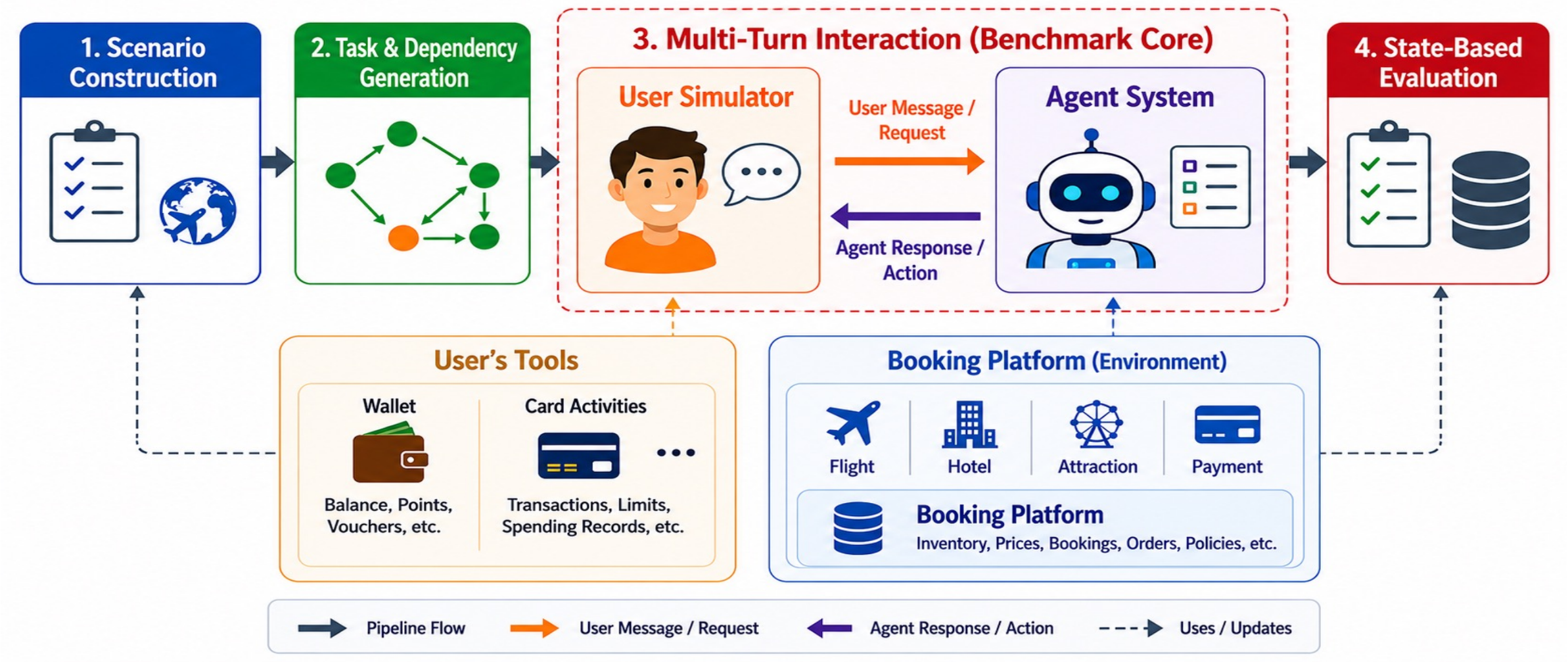}
  \caption{\bench Overview. User simulators interact with agent systems with their requests, and the agent uses diverse tools to solve the task. The final solution is verified based on the database state and the communication state. }
  \label{fig:overview}
\end{figure*}

\subsection{Task Formulation}

Similar to $\tau$-bench \cite{yao2025taubench}, the tasks in our benchmark have a three-component structure: a \textbf{user
simulator} $\mathcal{U}$, an \textbf{agent}, and an \textbf{environment} backed by a database $\mathcal{B}$ of the booking systems and a user private database of wallet and card $\mathcal{B}_{u}$. The agent is equipped with a tool set $\mathcal{T}_a$ and the user simulator with a tool set $\mathcal{T}_u$. As shown in \autoref{fig:overview}, the user simulators use $\mathcal{T}_u$ to interact with $\mathcal{B}_{u}$, and provide necessary information for booking. The agent books trips based on the user's request with $\mathcal{T}_a$ and modifies the booking system database $\mathcal{B}$.

\subsection{Constraint Graph-Based Task Generation}
\label{sec:task-gen}
As shown in \autoref{fig:task}, one trip booking task includes multiple subtasks, such as flight booking and hotel booking. Solving the task requires the agent to satisfy the user requirements in the entity (e.g.,  flight) in the subtask and resolve the dependencies between entities. 

However, it is challenging to initialize the booking system with multiple entities and their dependencies because: (i) there could be multiple valid solutions, making the task easy to solve; (ii) there could be no valid solution at all. Userbench~\cite{qian2025userbench}, which also creates travel planning tasks with multiple domains, solves these by using GPT-4o to generate about 100 combinations as options, and lets the agent choose the correct options. Such a setting reduces the difficulty in solving subtask dependencies, and cannot reflect how realistic trip booking is done. 

Instead, \bench creates independent databases for each entity and forces the agent to identify dependencies and determine the valid combinations of trips from the huge search space. To solve the above challenges, we propose a constraint-graph framework. It can construct an initial database state $\mathcal{B}_0$ that includes (i) at least one solution, (ii) multiple \emph{distractors}, which are confusing options that only partially satisfy entity or inter-entity constraints to mislead the agent.

\begin{figure}[ht]
  \centering
  \includegraphics[width=\linewidth]{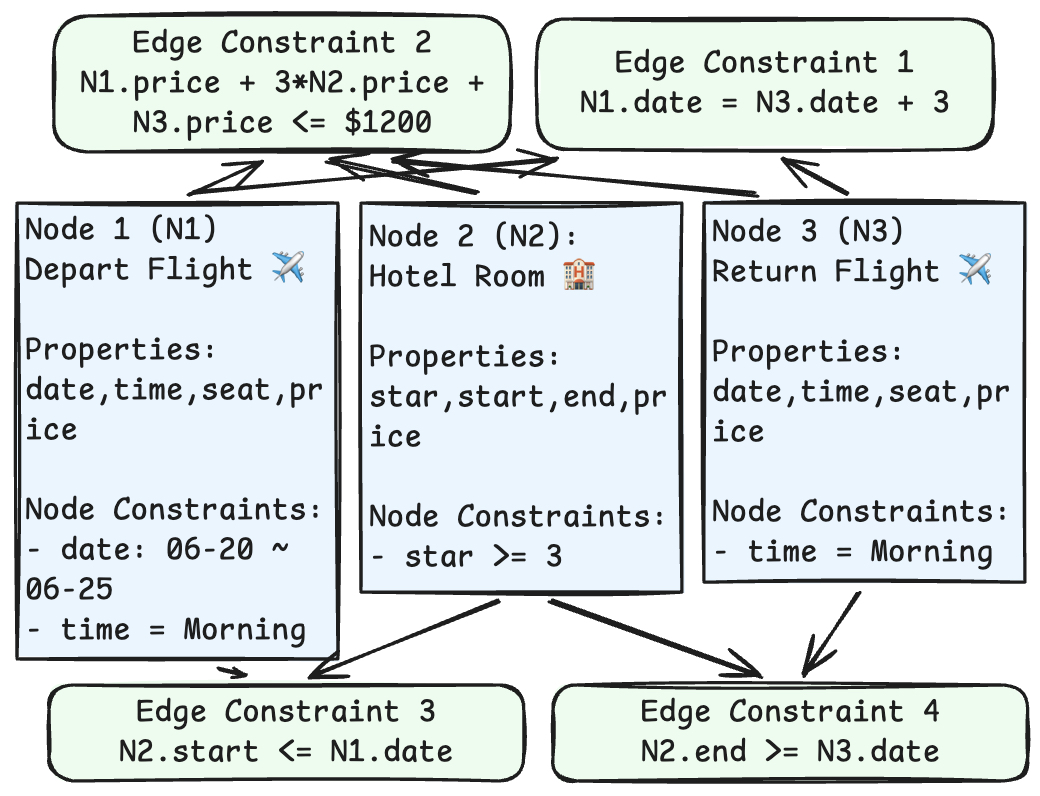}
  \caption{An example constraint graph. The user requirements are: A trip from Chicago to Pittsburgh. Flexible to leave between 06-20 and 06-25. Stay for 3 nights. Morning flights, a hotel $\geq$ 3-stars, budget of \$1200.}
  \label{fig:graph}
\end{figure}

\subsubsection{Nodes, Objects, and Constraints}

Each required entity in the task is represented as a
\textbf{node}.  A node defines a set of discrete-valued
\textbf{properties}, each with a finite domain of possible values.
For example, a \textsc{Flight} node has properties such as 
\texttt{departure\_time} $\in$ \{\textsc{Morning}, \textsc{Midday},
\textsc{Night}\}, \texttt{seat\_position} $\in$ \{\textsc{Window},
\textsc{Aisle}, \textsc{Middle}\}, \texttt{seat\_type} $\in$ \{\textsc{Economy},
\textsc{Business}\}, and etc. A
\textbf{concrete object} is a complete assignment of values to all
properties of a node.  A task graph
$G = (\mathcal{N}, \mathcal{C})$ collects all nodes
$\mathcal{N} = \{n_1, \ldots, n_k\}$ needed for the task and a set
of constraints $\mathcal{C}$ that specify which combinations of concrete objects constitute a valid solution.

Constraints in $\mathcal{C}$ fall into two classes.
\textbf{Domain constraints} are among the properties within a
specific domain. For example, a \textit{business-class} seat cannot have a \textit{middle} position. These constraints are shared across all tasks.
\textbf{Validity constraints} encode task-specific requirements related to the user's preferences, including node-level and edge-level. 
\textbf{Node-level} validity constraints restrict properties of a
single node. For example, in \autoref{fig:graph}, the user asks for a morning flight, so both Node 1 and Node 3 have $time=\text{Morning}$. 
\textbf{Edge-level} validity constraints involve
properties across two or more nodes and capture cross-object
coherence requirements. In \autoref{fig:graph}, the user's budget is \$1200, so the total price of three nodes should not exceed this budget (Edge Constraint 2). 
A valid solution is a tuple of concrete objects that satisfy all the constraints in $\mathcal{C}$.

\subsubsection{Seed Solution Generation}
To ensure there is at least one valid solution, we first generate several \textbf{seed solutions}, which are valid solutions that satisfy all constraints. When there are multiple seed solutions, cross-combinations of their objects may also form valid
solutions, as shown in the dotted lines in \autoref{fig:task}.

To generate a seed solution, we formulate a CSP over the properties referenced by the edge
constraints and sample a solution from it. With these values fixed,
each node is solved independently under its local domain and
node-level constraints, and one concrete object per node is sampled at random.

\subsubsection{Distractor Generation}

Based on the validity constraints, we generate two classes of distractors,
$D_{\text{node}}$ and $D_{\text{edge}}$, that inflate $\mathcal{B}_0$
without introducing additional valid solutions.

\paragraph{Node Distractors ($D_{\text{node}}$).}
For each node and each node-level validity constraint $c$, we
enumerate all realizable concrete objects that \emph{violate} $c$
while satisfying all domain constraints.  For example, for a task where the user requires a morning flight, this yields many flight objects at every
other time of day, and over all other combinations of properties which are individually well-formed but directly
fail a task requirement.  

\paragraph{Edge Distractors ($D_{\text{edge}}$).}
A valid flight may have no compatible hotel in the database: for instance, no hotel with a matching check-in date. We call such objects edge distractors: individually valid, but unable to form a complete solution with any other object.

We generate edge distractors as follows. For each edge constraint $c$, we maintain a pool of \emph{valid profiles}, which are the edge-relevant property values present in the seed solutions. A profile is \emph{universally bad} if it is incompatible with every profile currently in the pool from all other nodes. For example, a hotel at \$400/night is universally bad when every flight in the pool costs \$350 and the user's budget is \$700, since no valid pairing exists.

We iteratively generate edge distractor objects by choosing a universally bad profile, fixing its edge property values, and solve a per-node CSP to produce all concrete objects that satisfy the node's domain and node-level constraints. These objects are added to the pool as edge distractors. Then, we recompute which profiles remain universally bad and repeat until none are left. Details are in \ref{sec:graph_alg_app}.

\subsubsection{Database Initialization}

The output of the pipeline is a flat list of concrete objects tagged
as seed solutions, node distractor ($D_{\text{node}}$), or edge distractor ($D_{\text{edge}}$).  A
\textbf{instantiator} converts each object into one or more initialization actions that add the corresponding record into the database.
The full sequence of initialization actions executed on the database $\mathcal{B}$ produces the initial database $\mathcal{B}_0$ for the task. For the user database $\mathcal{B}_u$, we randomly initialize it with user and card information and ensure that at least one card has sufficient budget for the task. 

\subsection{User Simulation Grounded by Humans}
\label{sec:user}
We design 4 user behaviors and 3 user personas to narrow the gap between real-world human users and user simulators. Prompts are listed in \ref{sec:app_user}

\paragraph{User Behavior Dimensions.}
We follow the same behavior dimensions introduced in ~\citet{zhou2026mind} and to design 4 user behaviors for simulation: (D1) \texttt{Communication style}: how the user speak to the agents, such as their emotion, tone, the length of sentences, etc. (D2) \texttt{Information Disclosure}: how the user shares information; (D3) \texttt{Clarification}: how confident the user is when sharing information (D4) \texttt{Error Reaction}: how the user reacts to the agents' error. 

\paragraph{User Personas.}
We design 3 user personas that affect communication with agents: \texttt{Terse}, \texttt{Neutral}, and \texttt{Impatient}. \texttt{Terse} users communicate in short, clipped sentences, and get straight to the point with minimal information. \texttt{Impatient} users are less tolerant of too many questions and mistakes. \texttt{Neutral} users are the most collaborative ones that are kind. They always provide detailed information when asked. Each persona includes different choices of behavior dimensions. 
Different from $\tau^2-$ bench, which simulates users based on their age and background, our simulation focuses on human behavior patterns.

\subsection{State-based Evaluation}
\label{sec:eval}
There may be plenty of valid solutions that satisfy the user requirements in trip booking, making evaluation challenging. Therefore, instead of comparing the solution with the ground-truth set, we design a state-based evaluation, introducing the concrete state (e.g., booking system database $B$ and user private database $B_u$) and the abstract state (e.g., conversation). 
Concrete-state verification is a set of task-specific verifiers to test whether user requirements are satisfied in the final database. 
Abstract-state verification tests two properties: \textbf{factuality} (the agent's actions align with its stated plan, e.g.\ the booked flight ID and price match what was communicated) and \textbf{completion} (the interaction ends correctly, e.g.\ the agent offers a human transfer instead of leaving pending bookings unresolved). 

We view a task as solved only if the solution passes all verifiers. If the agent books a valid solution that differs from what it mentioned, it still fails because the solution cannot pass the factuality verifier. If the agent books a valid solution that differs from what it mentioned, it still fails because the solution cannot pass the factuality verifier.

\input{tab/benchmark}

\subsection{\bench}

We focus on trip booking because it involves multiple subtasks with dependencies. The 4 required entities in our benchmarks are \textit{departure flight, return flight, hotel, and attraction}. There are 6 types of edge constraints between nodes. The agent tool set $\mathcal{T}_a$ includes 19 tools across 5 categories, and the user tool set $\mathcal{T}_u$ includes 7 tools from 6 categories. The full set is in \autoref{tab:tool}.

With our automatic constraint graph-based task generation, given the properties, we can generate full combinations for all their possible values with random seed solutions. To keep our benchmark a reasonable size for evaluation, we create 4 sets with 1 to 4 seed solutions, generate full combinations for properties, and filter these tasks based on the process mentioned in Appendix \ref{sec:app_bench}. 

Finally, we keep 50 tasks for each seed solution group, naming them as S1, S2, S3, and S4, resulting in 200 tasks in total. S1 indicates the set of tasks starting from one valid seed solution. 
The detailed statistics are listed in \autoref{tab:benchmark}. 
With more seed solutions, the combinations of their different parts lead to more valid solutions. While the number of node distractors and edge distractors is similar, this increases the distractor ratio towards all possible solutions. This may make it easier for the agent to find the correct solutions. Therefore, we assume that S1 is the most difficult subset, while S4 is the easiest with the highest valid rate. 

Note that we can easily increase the number of tasks and extend it to other properties and domains based on our algorithm, making it easy to scale up.

\input{tab/main}

%% file: tab/benchmark.tex
\begin{table}[htbp]
  \centering
  \caption{\bench statistics. For each task, we have 4 nodes for 4 required entities and 6 edges for the task dependencies. Distractor ratio is $\frac{\#Valid}{|D_{node}| + |D_{edge}|}$, indicating difficulty in figuring out the solution. }
  \resizebox{\columnwidth}{!}{%
    \begin{tabular}{ccccc}
    \toprule
    \multicolumn{1}{l}{\# Seed} & \multicolumn{1}{l}{\# Valid} & \multicolumn{1}{l}{$|D_{node}|$} & \multicolumn{1}{l}{$|D_{edge}|$} & \multicolumn{1}{l}{Distractor Ratio}\\
    \midrule
    1  & 1.0  & 1819.3 & 214.8 & 0.05\% \\
    2  & 7.5 & 1819.3 & 204.3  & 0.37\%\\
    3  & 26.6 & 1781.2 & 226.1 & 1.32\%\\
    4  & 52.9 & 1780.0 & 228.9 & 2.62\%\\
    \bottomrule
    \end{tabular}%
    }
  \label{tab:benchmark}%
\end{table}%

%% file: tab/main.tex
\begin{table*}[h]
  \centering
  \caption{Results on \bench and the breakdown verification. One solution passes only if it passes both concrete and abstract-state verification. Generic indicates the default user simulator without the user persona and behaviors. $\Delta$ indicates the performance drop compared with the generic user simulator. }
  \resizebox{\textwidth}{!}{%
    \begin{tabular}{lccccccccc}
    \toprule
     & \multicolumn{3}{c}{Pass Rate} & \multicolumn{3}{c}{Concrete-state Verification} & \multicolumn{3}{c}{Abstract-state Verification} \\
    \cmidrule(lr){2-4} \cmidrule(lr){5-7} \cmidrule(lr){8-10}
     & \multicolumn{1}{l}{Generic} & \multicolumn{1}{l}{\user} & $\Delta$ & \multicolumn{1}{l}{Generic} & \multicolumn{1}{l}{\user} & $\Delta$ & \multicolumn{1}{l}{Generic} & \multicolumn{1}{l}{\user} & $\Delta$\\
    \midrule
    Claude Sonnet 4 6 & 0.52 & 0.32 & -38\% & 0.76 & 0.62 & -19\% & 0.62 & 0.60 & -3\% \\
    GLM-5 & 0.64 & 0.44 & -31\% & 0.76 & 0.56 & -27\% & 0.70 & 0.65 & -6\% \\
    Qwen3 Coder Next & 0.45 & 0.20 & -57\% & 0.47 & 0.28 & -39\% & 0.63 & 0.63 & 1\% \\
    DeepSeek V3.2 & 0.75 & 0.61 & -19\% & 0.88 & 0.73 & -17\% & 0.79 & 0.79 & 0\% \\
    \bottomrule
    \end{tabular}%
}
  \label{tab:main}%
\end{table*}%

%% file: sec/004evaluation.tex
In this section, we focus on two research questions: (i) how different agents perform on \bench? (ii) how \user affect the agent's performance?

\paragraph{Experimental Settings}
We follow the $\tau-$ benchmark framework~\citep{yao2025taubench,barres2025tau2} under its MIT License. 
We use \texttt{Claude Sonnet 4.6}, \texttt{GLM-5} (744B-A40B MoE), \texttt{Qwen3 Code Next} (80B-A3B MoE) and \texttt{DeepSeek V3.2}(685B-A37B MoE) as the agent. We use \texttt{Claude Sonnet 4.6} for the user simulation. 
We implement the human-aligned user simulator \user as described in Section \ref{sec:user} and use both the user persona (sampled uniformly) and all behavior dimensions. We also implement the generic user simulator, following $\tau^2-$bench.
We use the default inference and reasoning setting and report pass@1 in \autoref{tab:main} and discuss pass\^{}k in Section \ref{sec:app_more}.

\paragraph{DeepSeek V3.2 shows the best performance with 0.61, while Qwen3 Coder Next performs worst with 0.20.} Although DeepSeek V3.2 provide promising results, its pass@1 on \bench is much lower that on $\tau^2-$ bench (78.9\%). Similarly, Claude Sonnect 4.6 shows 79.5\% and GLM-5 gains 98.2\% on $\tau^2-$ bench Telecom domain~\footnote{https://benchlm.ai/benchmarks/tau2Bench}, but none of them get a pass@1 higher than 0.5 on \bench with our \user. This indicates \bench is more difficult,  benefiting from the constraints and dependencies in our benchmark.

\paragraph{Introducing human-like \user leads to consistent performance degradation across all agents.} The pass rate drops from $-19\%$ to $-57\%$. This indicates that while agents show a promising performance with generic user simulators, they cannot interact with human-like users reliably. This gap affects agents' capabilities in realistic scenarios. However, for a more powerful model (with a higher pass@1), such as \texttt{DeepSeek V3.2}, the influence is smaller. This indicates that further improving the agent's problem-solving capability may help it better interact with the human users.  

\paragraph{\user primarily degrades agents' ability to arrive at correct solutions, rather than their conversational quality.} Compared with the significant drop in concrete-state verification ($17$--$39\%$), which focuses on whether the agent can arrive at the correct solution, abstract-state verification is relatively stable  ($-6\%$ to $+1\%$). This indicates that human-like behaviors make the task more difficult for agents to solve, for example, by changing the way information is shared. However, these user behaviors have less impact on how agents interact with the user, such as how factual the agents are.

%% file: sec/005analysis.tex
\input{tab/breakdown}

\begin{figure}[ht]
    \centering
    \includegraphics[width=1\linewidth]{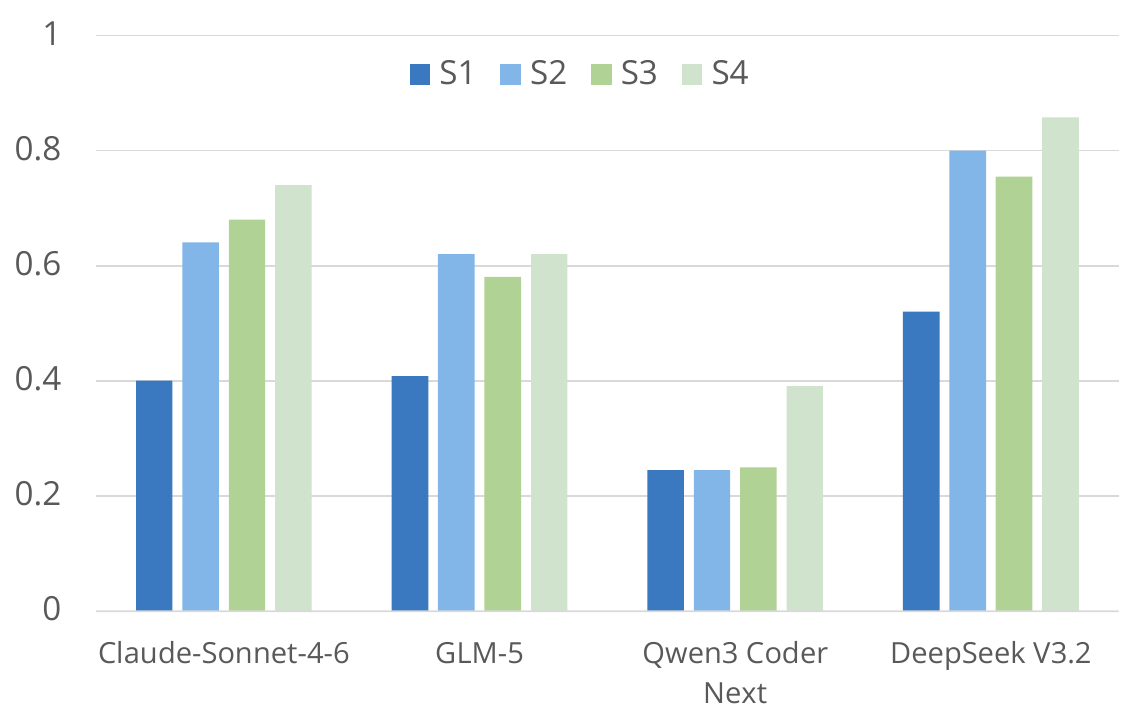}
    \caption{Concrete-state verification pass rate. S4 indicates 4 seed solutions. }
    \label{fig:ngt}
\end{figure}

\paragraph{Stronger models can better leverage alternative paths when available.}
We evaluate concrete-state performance across four task complexity levels, focusing on how the number of valid solutions affects the difficulty of arriving at the right database state.
As shown in \autoref{fig:ngt}, in general, the pass rate increases from S1 (hard) to S4 (easy).
However, compared with \texttt{DeepSeek V3.2} and \texttt{Claude Sonnet 4.6}, \texttt{Qwen3 Coder Next} does not benefit as much from the increase in valid solutions until the number of valid solutions becomes very large (S4). 
This indicates that stronger models are better at exploration of alternative paths when available, while weaker models remain bottlenecked.

\begin{figure}[ht]
    \centering
    \includegraphics[width=0.9\linewidth]{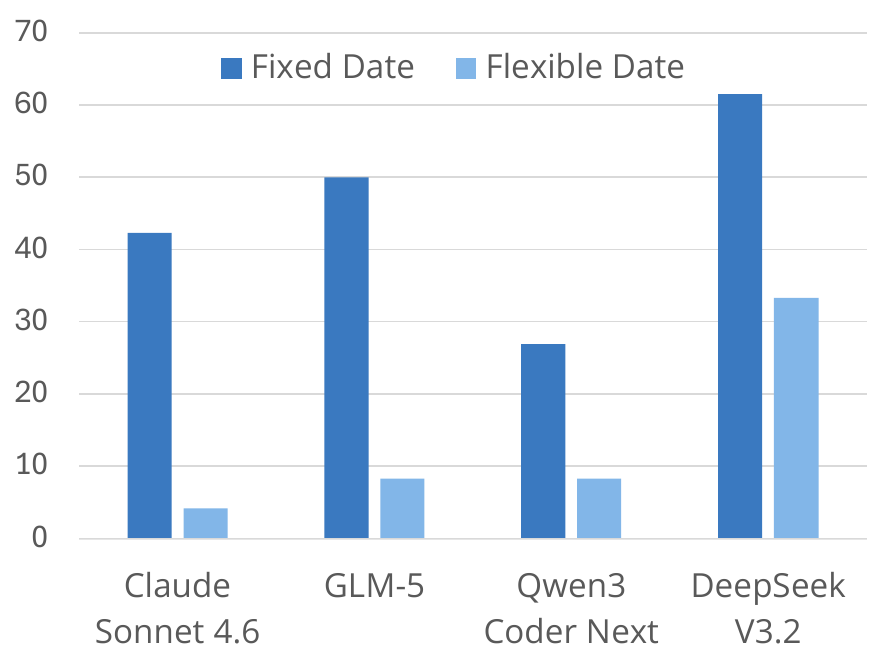}
    \caption{Pass@1 on fixed-date vs.\ flexible-date tasks. Flexible-date tasks have on average $5.5$ active edge constraints vs.\ $2.5$ for fixed-date tasks, with comparable distractor pool sizes.}
    \label{fig:flex}
\end{figure}

\paragraph{More edge constraints lead to substantially harder tasks, independent of distractor pool size.}
To investigate the effect of different edge constraints, 
we split \bench by date flexibility: whether the user's departure date is a single fixed day or can fall anywhere within a 5-day window. 
Both groups expose the agent to a comparable distractor pool (${\sim}2{,}066$ objects for fixed-date vs.\ ${\sim}1{,}999$ for flexible-date tasks). As shown in \autoref{fig:flex}, the difference in edge constraints alone produces large drops across all models: pass@1 on flexible-date tasks ranges from $8.3\%$ to $33.3\%$, compared to $26.9\%$--$61.5\%$ on fixed-date tasks. Even \texttt{DeepSeek V3.2}, the strongest model overall, achieves only $33.3\%$ on the flexible-date subset, isolating multiple subtask dependencies as a primary bottleneck for current agents.

\begin{figure*}[ht]
    \centering
    \includegraphics[width=1\linewidth]{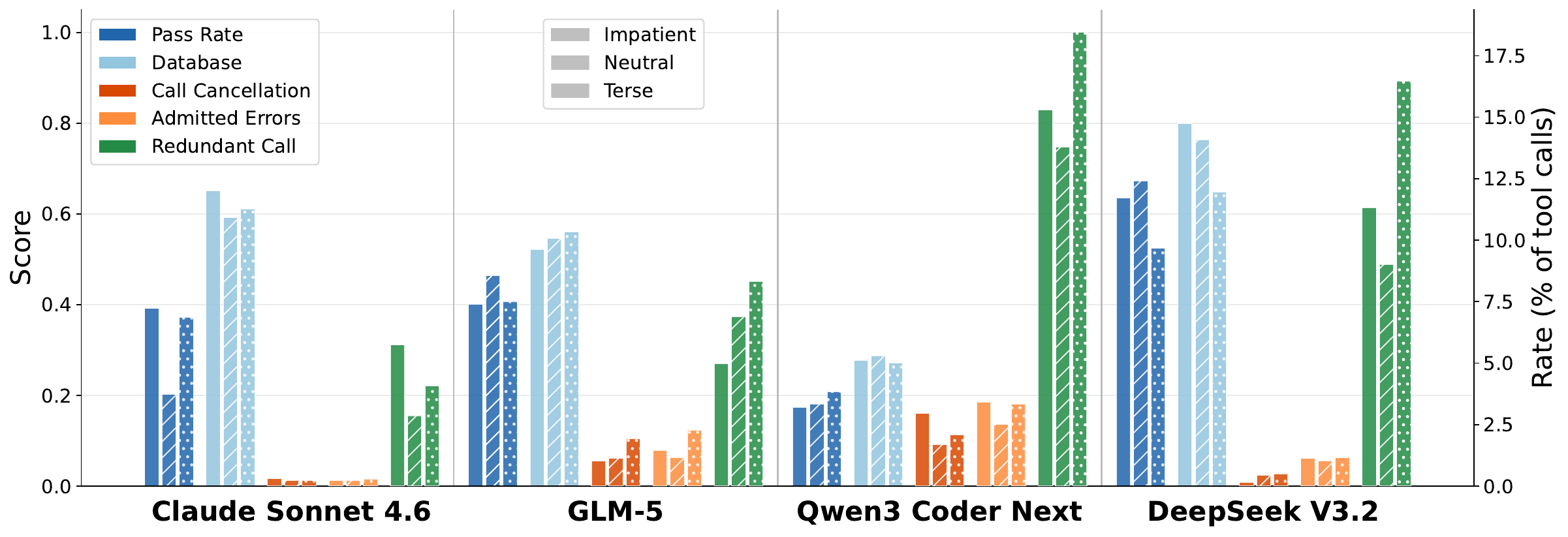}
    \caption{Performance analysis for different personas. The left y-axis is for Pass Rate and Database (concrete-state verification), and the right y-axis is for the rest. }
    \label{fig:persona}
\end{figure*}

\paragraph{Agents are less efficient towards \user, but they are less likely to admit mistakes.} 
We also investigate efficiency metrics to analyze the agent's performance, including the number of redundant calls (tool call with the same parameters), the number of cancellations during booking (e.g., the agent may book a wrong flight and then cancel), and the number of admitted errors (such as saying `\textit{I made a mistake}'). The detailed definitions are listed in Appendix \ref{sec:app_verifier}.
In \autoref{tab:breakdown}, factuality and completion are less affected by human-like behaviors, which is consistent with previous findings. 
However, the efficiency drops significantly among all agents. Redundant tool calls increase uniformly under the human-like condition, especially for \texttt{DeepSeek V3.2} and \texttt{Qwen3-Coder}. 
Call cancellation rates roughly double across all models. This indicates that, instead of asking clarification questions before booking, agents usually take action based on the underspecified requirement and then undo prior actions. 
Interestingly, while there are more call cancellations, admitted errors \emph{decrease} under the human-like condition for all models. This counterintuitive finding suggests that when interacting with more realistic users, agents are less likely to explicitly acknowledge mistakes. They are more likely to mask errors through implicit corrections or simply proceeding without disclosure, a behavior pattern that merits further investigation from a transparency perspective.

\begin{figure}[ht]
    \centering
    \includegraphics[width=1\linewidth]{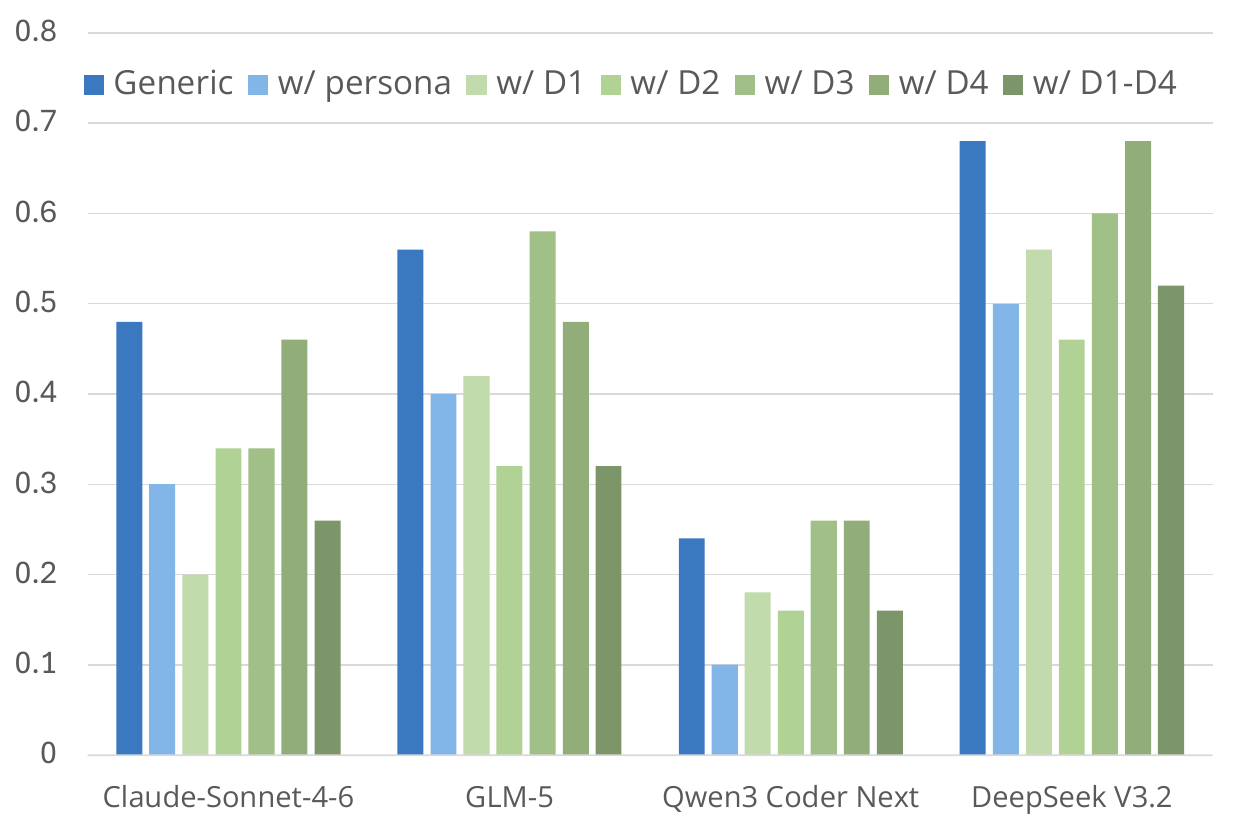}
    \caption{Pass@1 of different types of user simulation. Results are from \texttt{Claude Sonnect 4.6} on the S1 subset. }
    \label{fig:ablation}
\end{figure}

\paragraph{Information Disclosure (D2) is the most challenging behavior.}
We isolate the contribution of each behavioral dimension and persona to the overall difficulty introduced by the human-like user simulator. 
Starting from the generic user, we independently add persona descriptions and individual behavioral dimensions D1--D4. As shown in \autoref{fig:ablation}, assigning a persona alone causes substantial drops across all models. 
Poor performance in \textit{Information Disclosure} indicates that how users expose information to the agent significantly affects the agent's capabilities. D1 (\textit{Communication Style}) is most damaging to Claude Sonnet ($0.48 \to 0.20$), while D3 (\textit{Clarification}) and D4 (\textit{Error Reaction}) have comparatively mild effects.

\paragraph{Persona affects agent efficiency more than task success, and \texttt{Terse} significantly degrades efficiency.} 
As shown in \autoref{fig:persona}, \texttt{Terse}, who provide minimal information, increases the redundant tool calls and admitted errors, compared with \texttt{Natural}. The agents are forced to probe the environment when they cannot confirm details through conversation. 
\texttt{GLM-5} doubles its Call Cancellation with \texttt{Terse} vs the other two. This indicates that without user elaboration, it books prematurely, then corrects, making booking inefficient.
Despite expressing frustration, \texttt{Impatient} users don't degrade task success. They actually signal problems clearly (e.g., `you made an error'), which may help the agent recover.

\paragraph{Failure modes reveal systematic gaps in tool use, constraint reasoning, and transparency.}
\autoref{tab:failure_db} and \autoref{tab:failure_comm} summarize common failure modes. Among concrete-state failures, \textbf{payment inability} dominates, occurring when the agent fails to use payment tools correctly. The second most common failure reflects agents' inability to resolve inter-entity constraints. Among factuality failures, the dominant pattern is agents booking hotels or attractions \textit{without first disclosing the name to the user} (\texttt{Claude Sonnet 4.6} accounts for 84 such cases), followed by promise-action mismatches where agents state one plan but execute another (ID mismatch: 11\%; plan deviation: 12\%). For completion failures, 96\% are from agents leaving incorrect bookings uncancelled without notifying the user. Together, these failures point to three directions for improvement: better payment tool grounding, stronger inter-entity constraint propagation, and transparency mechanisms enforcing consistency between agent statements and actions.

\input{tab/failure_db}
\input{tab/failure_comm}

%% file: tab/breakdown.tex
\begin{table*}[ht]
  \centering
  \caption{Communication and efficiency performance on \bench. $\uparrow$ indicates the higher the better. }
  \resizebox{\textwidth}{!}{%
    \begin{tabular}{lcccccccccc}
    \toprule
       & \multicolumn{4}{c}{Communication} & \multicolumn{6}{c}{Efficiency } \\
       \cmidrule(lr){2-5} \cmidrule(lr){6-11} 
       & \multicolumn{2}{c}{Factuality $\uparrow$} & \multicolumn{2}{c}{Completion $\uparrow$} & \multicolumn{2}{c}{Redundant Calls $\downarrow$ } & \multicolumn{2}{c}{Call Cancellation $\downarrow$} & \multicolumn{2}{c}{Admitted Errors $\downarrow$} \\
       \cmidrule(lr){2-3}  \cmidrule(lr){4-5} \cmidrule(lr){6-7} \cmidrule(lr){8-9} \cmidrule(lr){10-11} 
     & Generic & \user & Generic & \user & Generic & \user & Generic & \user & Generic & \user \\
    \midrule
    Claude Sonnet 4.6 & 89.6\% & 86.6\% & 95.8\% & 96.3\% & 3.4\% & 4.2\% & 0.1\% & 0.2\% & 2.7\% & 0.3\% \\
    GLM-5 & 91.7\% & 88.7\% & 94.5\% & 92.7\% & 3.8\% & 6.9\% & 0.6\% & 1.4\% & 3.1\% & 1.6\% \\
    Qwen3 Coder Next & 88.1\% & 91.1\% & 93.7\% & 89.3\% & 12.8\% & 15.8\% & 0.6\% & 2.1\% & 8.2\% & 3.1\% \\
    DeepSeek V3.2 & 97.0\% & 96.4\% & 96.0\% & 95.7\% & 6.6\% & 12.6\% & 0.2\% & 0.4\% & 4.6\% & 1.1\% \\
    \bottomrule
    \end{tabular}%
}
  \label{tab:breakdown}%
\end{table*}%

%% file: tab/failure_db.tex
\begin{table}[ht]
\centering
\caption{Concrete-state failure modes (count=356).}
\resizebox{\columnwidth}{!}{%
\begin{tabular}{llr}
\toprule
\textbf{Failure Mode} & \textbf{Key Insight}\\
\midrule
Payment inability (37\%) & Fails to handle payment  \\
Wrong items booked (23\%) & Trip violates requirements \\
Compromised users (25\%) & User gives up early \\
Partial booking (6\%) & Trip only partially fulfilled  \\
\bottomrule
\end{tabular}%
}
\label{tab:failure_db}
\end{table}

%% file: tab/failure_comm.tex
\begin{table}[ht]
\centering
\caption{Abstract-state failure modes. Factuality: 280; Completion: 102.}
\resizebox{\columnwidth}{!}{%
\begin{tabular}{ll}
\toprule
\textbf{Category} & \textbf{Failure Mode} \\
\midrule
\multirow{5}{*}{Factuality}
  & Name not disclosed (56\%)  \\
  & Summary amount $\neq$ actual charge (16\%) \\
  & Booked items $\neq$ approved plan (12\%) \\
  & ID mismatch (stated vs.\ booked) (11\%) \\
  & Price mismatch (item-level or pre-charge) (4\%) \\
\midrule
Completion & No transfer offered for partial booking (96\%) \\
\bottomrule
\end{tabular}%
}
\label{tab:failure_comm}
\end{table}

%% file: sec/006conclusion.tex
We presented \bench and \user, an agentic evaluation framework designed to expose the gap between benchmark performance and real-world agent capability.
\bench generates tasks with complex inter-entity dependencies and structured distractors, while \user replaces idealized user simulators with behavior grounded in human studies.
Together, they reveal that frontier LLM agents, despite near-perfect scores on existing benchmarks, struggle substantially when faced with realistic constraints and imperfect users.
Our analysis shows that the performance gap is not about conversation quality, which remains largely intact, but about the agent's ability to reason over constrained solution spaces under ambiguous and incremental information disclosure. \bench and \user can serve as a foundation for developing agents that are robust not only to hard tasks, but to the full complexity of real human interaction.

%% file: sec/020limitation.tex
\bench is currently instantiated in the trip booking domain, which, while rich in subtask dependencies, may not capture failure modes specific to other agentic settings such as coding assistance. Extending the constraint graph framework to new domains only requires defining entity schemas and constraint types, making it easy to instantiate in new domains.
\user covers four behavioral dimensions grounded in a prior human study, but real user behavior is broader and more variable. Dimensions such as deception, topic drift, or highly domain-specific jargon are not modeled. In addition, although \user is grounded in behavioral observations from human studies, the simulators are still LLM-based and inherit the calibration limitations documented in prior work~\citep{seshadri2026lost,zhou2026mind}.

%% file: sec/010appendix.tex
\subsection{Verifiers and Metrics}
\label{sec:app_verifier}
We use concrete-state verification to check the final state of the database, and use abstract-state verification to check the communication quality. One solution needs to pass both verifications to be correct.

Abstract-state verification includes \textit{Factuality} and \textit{Completion}. Factuality verifies whether the agent's actions align with what they say, and Completion verifies whether the agent terminates the conversation in the right way. Both types of verifiers are rule-based and only have the outcome of True or False.

\paragraph{Factuality verifiers.} We inspect the factuality from the following aspects. Note that the verifier is only triggered when the conversation mentions the specific item. For example, if the agent does not provide a post-booking summary, the corresponding verifier is omitted. 

\begin{itemize}[topsep=0pt,itemsep=2pt,parsep=0pt,leftmargin=1.5em]
    \item \textbf{Pre-charge totals}: the dollar amounts stated before making a charge must match the actual charges 
    \item \textbf{Post-booking summary}: the final total must match the confirmed bookings of the database
    \item \textbf{Booking IDs}: flight/hotel/attraction IDs passed to book\_* tools must have been mentioned to the user beforehand
    \item \textbf{Booking names}: hotel/attraction names must have been communicated before booking
    \item \textbf{Item-level prices}: per-item prices stated by the agent must match actual charges
    \item \textbf{Approved plan match}: items actually booked must match the IDs the user explicitly approves
\end{itemize}

\paragraph{Completion verifiers.}
Following the $\tau^2-$bench, we have three final conversation states: STOP, TRANSFER, OUT-OF-SCOPE. OUT-OF-SCOPE is reserved because we ensure that all tasks are solvable with the given tools and initial database. Therefore, Completion verifiers include: 

\begin{itemize}[topsep=0pt,itemsep=2pt,parsep=0pt,leftmargin=1.5em]
    \item Never end with OUT-OF-SCOPE
    \item If the trip is fully booked with all required entities (flights, hotel, attractions), it should end with STOP (successfully complete) or TRANSFER (fail and need help from human agents).
    \item If the trip is partially booked, the agent must offer a transfer to a human agent
    \item If no booking at all, the conversation should end with STOP (e.g., the user gives up).
\end{itemize}

\paragraph{Efficiency Metrics}
Furthermore, we investigate the efficiency of communication from tool calls and agent mistakes. Note that these metrics are only for analysis purposes and will not affect the pass rate. 

\begin{itemize}[topsep=0pt,itemsep=2pt,parsep=0pt,leftmargin=1.5em]
    \item Total tool calls: all tool calls made by the agent
    \item Call Cancellation: Unnecessary booking changes, such as book an incorrect flight and then canceling it.
    \item Admitted errors: messages mention "I apologize for the error", "my mistake", etc.
    \item Failed Tool Calls: tool call fails
    \item Redundant Calls: exact-duplicate calls with the same name and arguments. 
\end{itemize}

\subsection{User Simulation}
\label{sec:app_user}
\begin{LLMPrompt}[title={User Persona Prompt}]
Here are the 3 personas and 4 behavior dimensions we used in \bench.

\texttt{Terse} \\
You are a no-nonsense person who communicates in short, clipped sentences. You rarely use pleasantries and prefer to get straight to the point. You answer questions with the minimum information needed and do not elaborate unless pressed. \\
 \\
\texttt{Neutral} \\
You are a straightforward person. You are polite but not overly friendly. You provide information when asked without much embellishment. Your tone is conversational but efficient. \\
 \\
\texttt{Impatient} \\
You are in a hurry and slightly irritable. You want the booking done quickly. If the agent asks too many questions or makes errors, you become noticeably curt. You do not apologize for the agent's mistakes and you express brief frustration when things go wrong.

\end{LLMPrompt}

\begin{LLMPrompt}[title={Behavior Demension Prompt}]

\textbf{(D1) Communication style}: keep your replies short — fragments and single sentences are fine. Do not pad messages with pleasantries like 'Sure!', 'Thank you so much!', or 'That sounds great!'. Vary your tone naturally; you may be curt, neutral, or brief. Never open with a greeting longer than one short sentence. \\

\textbf{(D2) Information disclosure}: only share information from the 'Say upfront' section in your first message. Reveal other details only when the agent asks for them or when they become relevant. Do not volunteer all your requirements at once. \\

\textbf{(D3) Clarification}: state your preferences plainly when asked. Do not stack multiple hedges or dramatic uncertainty markers (e.g., 'I'm crossing my fingers', 'maybe I'm cursed'). At most one brief hedge per turn is acceptable (e.g., 'I'm not sure about the dates'). \\

\textbf{(D4) Error reactions}: if the agent makes a mistake or retrieves wrong information, correct it tersely (e.g., 'No, wrong dates.' or 'That's not it.'). Do not apologize for the agent's error, do not cooperatively pivot to a different strategy, and do not volunteer alternative information to help the agent recover. If the agent repeats the same mistake, you may express brief frustration. \\
\end{LLMPrompt}

\subsection{Benchmark details}
\label{sec:app_bench}
\paragraph{Information Control for User Simulation}
To control how the user simulator provides information to the agent, we split the user's known information into two parts: one set of information is the basic information, which can be said upfront; the other set of information is detailed information, which should only be provided when asked. For example, as shown in \autoref{fig:task}, when booking a trip, the departure day and the duration can be provided in the first round. The other information, such as the name, the traveler's date of birth, and the traveling preferences, will not be revealed until the agent explicitly asks. We use this two-level information sharing as the default behavior in the user personas setting. In the behavior dimensions setting, we implement this for \texttt{Information Disclosure}.

\begin{figure}[ht]
    \centering
    \includegraphics[width=1\linewidth]{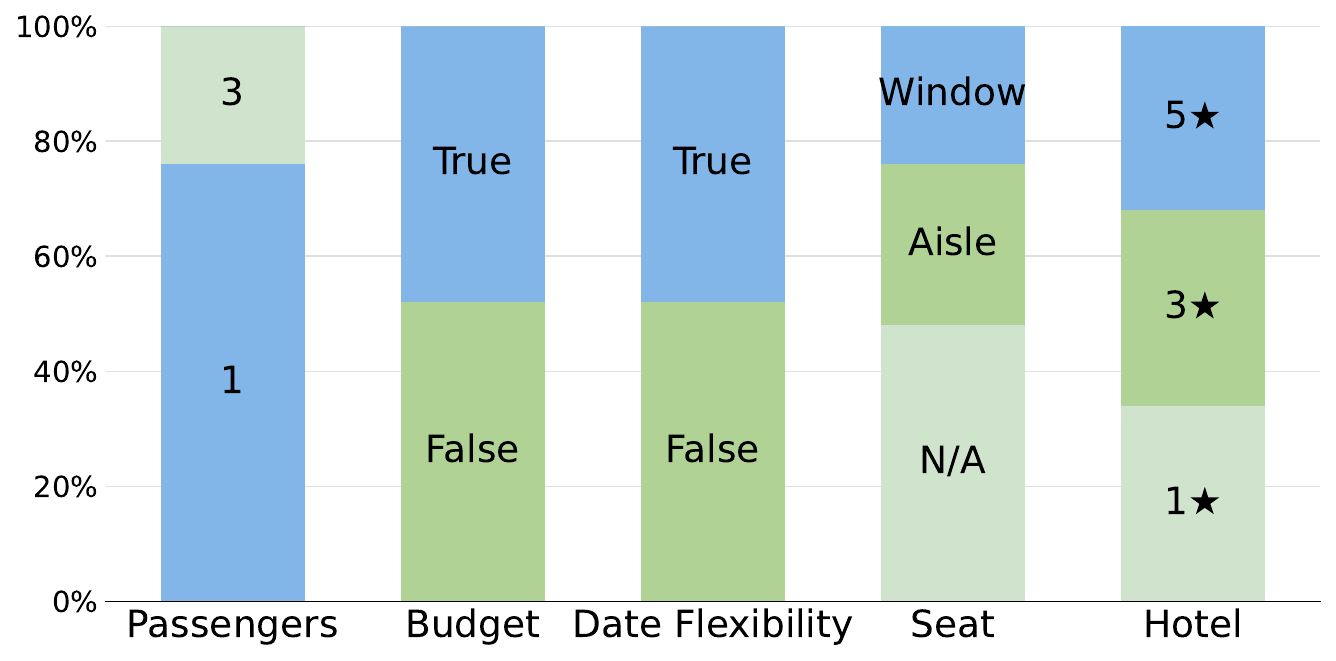}
    \caption{Distribution of the important properties to ensure the diversity of our benchmark. }
    \label{fig:bench}
\end{figure}

\paragraph{Diversity and Complexity Balance in \bench}
After creation, we first de-duplicate the seed solutions and then sort the tasks based on the number of valid solutions, $D_{node}$ and $D_{edge}$. Typically, the task with fewer valid solutions and more distractors requires the agent to search more, making it more difficult. Finally, we use a round-robin across strata to balance the difficulty and diversity. We pick the next-hardest task in each group in turn. The groups are split based on important properties that will affect booking, such as the number of passengers, the budget, the date flexibility, the seat preference, and hotel star ratings. the distribution of properties is shown in \autoref{fig:bench}.

\paragraph{Tool Definition}
We list all tools we use in \autoref{tab:tool}. For trip booking, we also implement the modification and cancellation tools in order to recover from incorrect bookings.

\input{tab/tools}

\subsection{Graph Algorithm Details}
\label{sec:graph_alg_app}

\paragraph{Notation.}
Let $G = (\mathcal{N}, \mathcal{C})$ be a task constraint graph.
Each \textbf{node} $n \in \mathcal{N}$ represents a required entity
type (e.g.\ a flight seat or a hotel room) and is defined by a set
of discrete-valued properties, each with a finite domain $D(p)$.
The constraint set $\mathcal{C}$ is split into two disjoint subsets:
\begin{itemize}[nosep,leftmargin=*]
  \item $\mathcal{C}_{\text{node}}$: \textbf{node constraints}, predicates
        over properties of a \emph{single} node encoding per-entity
        task requirements (e.g.\ departure time must be morning).
  \item $\mathcal{C}_{\text{edge}}$: \textbf{edge constraints}, predicates
        over properties drawn from \emph{two or more} nodes encoding
        cross-entity coherence requirements
        (e.g.\ $\texttt{flight.price} + \texttt{hotel.price} \leq B$).
\end{itemize}
An \textbf{assignment} for node $n$ is a mapping from each of $n$'s
properties to a value in its domain, satisfying $n$'s domain
constraints.  A \textbf{valid solution} is a tuple of assignments,
one per node, that jointly satisfy every constraint in $\mathcal{C}$.
The \textbf{profile} of an assignment $o$ with respect to an edge
constraint $c$, written $\text{profile}(o, c)$, is the sub-tuple of
values for the properties of $o$'s node that are referenced by $c$.

The overall goal of the generation pipeline is to produce a large
list of assignments (both seed assignments from valid solutions, and
distractors) that are subsequently \emph{materialized} into the
initial database $\mathcal{B}_0$ before dialogue begins
(see Section~\ref{sec:materialization}).

\medskip
\paragraph{Helper functions.}
The algorithms below rely on three abstract helpers:
\begin{itemize}[nosep,leftmargin=*]
  \item $\text{SampleCSP}(\mathcal{V}, \mathcal{C}', k)$: return up
        to $k$ randomly sampled solutions from the CSP defined over
        variable set $\mathcal{V}$ with constraints $\mathcal{C}'$.\footnote{%
          Implemented via \texttt{python-constraint} (backtracking
          solver).  Two strategies are supported: enumerate all
          solutions and draw $k$ at random, or shuffle variable
          domains before each call and extract the first solution,
          avoiding full enumeration when the solution space is large.}
  \item $\text{SolveNode}(n, \mathcal{C}', \mathit{ov})$: return all
        assignments for node $n$ satisfying its domain
        constraints, the node constraints $\mathcal{C}'$, and any
        property values pinned by overrides $\mathit{ov}$.
  \item $\text{SolveGraph}(G, k)$: return up to $k$ solutions from
        the full graph CSP (used only as a fallback in
        Algorithm~\ref{alg:gt}).
\end{itemize}

\noindent\textbf{Subroutine} \textsc{UniversallyBad}$(\pi, n, c)$:
returns \texttt{true} iff profile $\pi$ for node $n$ violates $c$
for \emph{every} combination of currently valid profiles from all
other nodes involved in $c$:
\begin{align*}
  &\forall\; n' \neq n,\; \pi_{n'} \in \mathit{valid}[n'] :\\
  &\quad \text{assemble}\!\left(\pi,\, \{\pi_{n'}\},\, c\right) \in \mathcal{I}_c
\end{align*}

\subsection{Materialization}
\label{sec:materialization}

Once the generation pipeline produces its full list of assignments
(seed assignments and distractors), each must be converted into one
or more concrete database insertion calls before dialogue begins.
This conversion is handled by a \textbf{materializer}, which is
type-driven: the node type of each assignment selects the appropriate
insertion routine, and the property values are translated into the
arguments of the corresponding tool call.

For the travel domain, there are three node types.

\paragraph{Flight seat assignments (\texttt{flight\_seat\_set}).}
Each assignment specifies a route (origin, destination, duration), a
departure date, a time of day, a seat type (economy, premium economy,
or business), a seat position (window, aisle, or middle), and a price
level.  Before insertion, all assignments sharing the same (route,
departure date, time of day) triplet are grouped into a single
physical flight, since they represent different seat configurations
on the same service.  The materializer issues one
\texttt{add\_flight\_with\_seats} call per group, passing the route
and date once and bundling all seat specifications (type, position,
count, and mapped dollar price) as a single list argument.

\paragraph{Hotel room assignments (\texttt{hotel\_room\_availability}).}
Each assignment specifies a city, a star rating, a room type, a bed
type, a maximum occupancy, an availability start date, a number of
available nights (\texttt{num\_dates}), and a price per night.
Assignments sharing the same (city, star rating) pair are grouped
into one hotel entity.  The materializer issues one
\texttt{add\_hotel} call per group, then one
\texttt{add\_room\_to\_hotel} call per assignment, setting the
availability window to
$[\texttt{start\_date},\; \texttt{start\_date} + \texttt{num\_dates}]$.

\paragraph{Attraction assignments (\texttt{attraction\_slot}).}
Each assignment specifies a city, a category (museum, tour, or show),
a date, a time of day (morning, afternoon, evening, or all-day), and
a price per ticket.  Each assignment maps one-to-one to an
\texttt{add\_attraction} call with no grouping.  Time-of-day values
are translated to a concrete start/end window (e.g.\
\textsc{Morning} $\to$ 09:00--12:00); all-day events receive a
00:00--23:59 window.

\input{algo/algo1}

\input{algo/algo2}

\input{algo/algo3}

\subsection{More Experiment Results}
\label{sec:app_more}
\paragraph{Huge gap between pass@k and pass\^{}k indicates models still struggle with solving tasks reliably.}
With 4 independent trials per task, we report pass@k (at least one success in $k$ trials) and pass\^{}k (all $k$ trials succeed) to measure both capability coverage and behavioral consistency. \texttt{DeepSeek V3.2} achieves the highest pass@4, demonstrating that the majority of tasks are within its capability frontier, yet its pass\^{}4 of 0.12 reveals that only a small fraction of tasks are reliably solved. This gap between coverage and consistency is shared across all models: 
\texttt{Qwen3 Coder Next} exhibits the lowest performance on both metrics.

\begin{figure}[ht]
    \centering
    \includegraphics[width=1\linewidth]{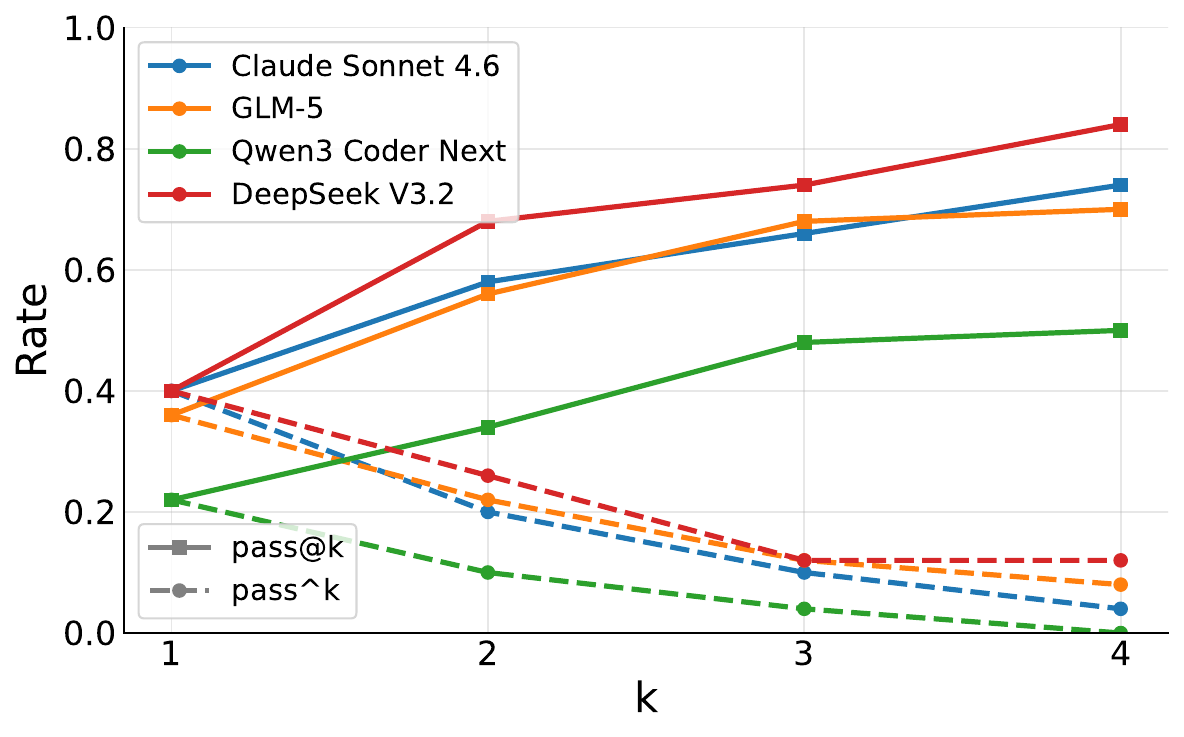}
    \caption{Pass\^{}k and pass@k for different agents. }
    \label{fig:pass}
\end{figure}

\subsection{Potential Risks}
The failure modes identified in our analysis, particularly agents booking items that contradict what they told the user, highlight transparency risks in deployed systems. Benchmarks that do not evaluate factuality between agent speech and action may miss this class of failures entirely, and we encourage future work to treat communication consistency as a first-class evaluation criterion.

\subsection{The Use of LLMs}
A large language model (LLM) was used as a general-purpose assistive tool to check grammar and correct typographical errors in this paper. The LLM did not contribute to research ideation, experimental design, analysis, or substantive writing. The authors take full responsibility for the content of the paper.

%% file: tab/tools.tex
\begin{table*}[t]
  \centering
  \caption{Tool usage in the Travel domain simulation. R = Read, W = Write, G = Generic.}
  \label{tab:tool}
  \small
  \begin{tabular}{llll}
  \toprule
  \textbf{Tool Name} & \textbf{Owner} & \textbf{Type} & \textbf{Category} \\
  \midrule
  \texttt{get\_customer\_information} & Agent & R & Platform \\
  \texttt{update\_customer} & Agent & W & Platform \\
  \texttt{transfer\_to\_human\_agents} & Agent & G & Platform \\
  \texttt{list\_all\_airports} & Agent & R & Flights \\
  \texttt{search\_flights\_by\_route} & Agent & R & Flights \\
  \texttt{get\_flight\_booking\_details} & Agent & R & Flights \\
  \texttt{get\_price\_airline\_booking} & Agent & R & Flights \\
  \texttt{search\_available\_seats} & Agent & R & Flights \\
  \texttt{book\_flight\_with\_seats} & Agent & W & Flights \\
  \texttt{cancel\_flight} & Agent & W & Flights \\
  \texttt{search\_hotels\_by\_city} & Agent & R & Hotels \\
  \texttt{search\_available\_rooms} & Agent & R & Hotels \\
  \texttt{get\_price\_hotel\_booking} & Agent & R & Hotels \\
  \texttt{book\_hotel\_with\_rooms} & Agent & W & Hotels \\
  \texttt{search\_attractions\_by\_city} & Agent & R & Attractions \\
  \texttt{book\_attraction} & Agent & W & Attractions \\
  \texttt{get\_recent\_payment\_transactions} & Agent & R & Payments \\
  \texttt{get\_transaction\_details} & Agent & R & Payments \\
  \texttt{charge\_booking} & Agent & W & Payments \\
  \midrule
  \texttt{get\_my\_payment\_cards} & User & R & Wallet \\
  \texttt{set\_default\_payment\_card} & User & W & Wallet \\
  \texttt{get\_my\_trip\_confirmations} & User & R & Confirmations \\
  \texttt{get\_trip\_spending\_summary} & User & R & Spending \\
  \texttt{get\_recent\_card\_activity} & User & R & Card Activity \\
  \texttt{record\_payment\_approval} & User & W & Approvals \\
  \texttt{add\_payment\_method\_to\_platform} & User & W & Platform \\
  \bottomrule
  \end{tabular}
\end{table*}

%% file: algo/algo1.tex
\begin{algorithm}[htbp]
\caption{GenerateSeedSolutions($G$, $k$, $m$)}
\label{alg:gt}
\begin{algorithmic}[1]
\Require Graph $G{=}(\mathcal{N},\mathcal{C})$,
         desired seed count $k$,
         attempt multiplier $m$
\Ensure  Set of seed solutions $\mathcal{S}$
\State $\mathcal{S} \leftarrow \emptyset$
\State $\mathcal{V} \leftarrow$ properties referenced by any $c \in \mathcal{C}_{\text{edge}}$
\State $\mathcal{C}_{\text{mini}} \leftarrow
       \{c \in \mathcal{C} \mid \text{all variables of } c \text{ are in } \mathcal{V}\}$
\State $B \leftarrow \text{SampleCSP}(\mathcal{V},\; \mathcal{C}_{\text{mini}},\; m{\cdot}k)$
\For{each binding $\beta \in B$}
    \State $\mathit{sol} \leftarrow \{\}$
    \State $\mathit{ok} \leftarrow \texttt{true}$
    \For{each node $n \in \mathcal{N}$}
        \State $\mathit{ov} \leftarrow$ values in $\beta$ for properties of $n$
        \State $O \leftarrow \text{SolveNode}(n,\; \mathcal{C}_{\text{node}}(n),\; \mathit{ov})$
        \If{$O = \emptyset$}
            \State $\mathit{ok} \leftarrow \texttt{false}$;\enspace \textbf{break}
        \EndIf
        \State $\mathit{sol}[n] \leftarrow \text{UniformSample}(O)$
    \EndFor
    \If{$\mathit{ok}$ \textbf{and} $\mathit{sol} \notin \mathcal{S}$}
        \State $\mathcal{S} \leftarrow \mathcal{S} \cup \{\mathit{sol}\}$
    \EndIf
    \If{$|\mathcal{S}| = k$}\enspace \textbf{break} \EndIf
\EndFor
\If{$|\mathcal{S}| < k$}
    \State $\mathcal{S} \leftarrow \text{SolveGraph}(G, k)$
    \Statex \hspace{2.5em}\textit{// fallback: full graph CSP}
\EndIf
\State \Return $\mathcal{S}$
\end{algorithmic}
\end{algorithm}

%% file: algo/algo2.tex
\begin{algorithm}[htbp]
\caption{GenerateNodeDistractors($G$, $\rho$)}
\label{alg:node-dist}
\begin{algorithmic}[1]
\Require Graph $G{=}(\mathcal{N},\mathcal{C})$,
         sampling ratio $\rho \in (0,1]$
\Ensure  Node distractor set $D_{\text{node}}$
\State $D_{\text{node}} \leftarrow \emptyset$
\For{each node $n \in \mathcal{N}$}
    \For{each $c \in \mathcal{C}_{\text{node}}(n)$}
        \State $\mathcal{C}^{-c} \leftarrow
               \bigl(\mathcal{C}_{\text{node}}(n) \setminus \{c\}\bigr) \cup \{\neg c\}$
        \Statex \hspace{3em}\textit{// all node constraints, with $c$ negated}
        \State $V \leftarrow \text{SolveNode}(n,\; \mathcal{C}^{-c},\; \{\})$
        \Statex \hspace{3em}\textit{// well-formed objects that violate $c$}
        \State $D_{\text{node}} \leftarrow D_{\text{node}} \cup
               \text{UniformSample}(V,\; \lceil\rho \cdot |V|\rceil)$
    \EndFor
\EndFor
\State \Return $D_{\text{node}}$
\end{algorithmic}
\end{algorithm}

%% file: algo/algo3.tex
\begin{algorithm}[htbp]
\caption{GenerateEdgeDistractors($G$, $\mathcal{S}$)}
\label{alg:edge-dist}
\begin{algorithmic}[1]
\Require Graph $G{=}(\mathcal{N},\mathcal{C})$,
         seed solutions $\mathcal{S}$
\Ensure  Edge distractor set $D_{\text{edge}}$
\State $D_{\text{edge}} \leftarrow \emptyset$
\For{each $c \in \mathcal{C}_{\text{edge}}$}
    \State $\mathcal{I}_c \leftarrow
           \bigl\{\mathbf{t} \in \prod_{(n,p)\in c} D(p) \;\big|\; c(\mathbf{t}) = 0\bigr\}$
    \Statex \hspace{2.5em}\textit{// all value tuples violating $c$}
    \For{each node $n$ in $c$}
        \State $\mathit{valid}[n] \leftarrow
               \{\text{profile}(o, c) \mid o \in \text{seed objects of } n\}$
        \Statex \hspace{4.5em}\textit{// init from seed objects}
        \State $P[n] \leftarrow \text{SolveNode}(n,\;\mathcal{C}_{\text{node}}(n),\;\{\})$
        \Statex \hspace{4.5em}\textit{// all node-valid objects}
        \State $\mathit{bad}[n] \leftarrow \{\,\text{profile}(o,c) \mid o \in P[n],$
        \Statex \hspace{4.5em}$\text{UniversallyBad}(\text{profile}(o,c),\, n,\, c)\,\}$
    \EndFor
    \While{$\exists\, n: \mathit{bad}[n] \neq \emptyset$}
        \State Pick $(n, \pi)$ with $\pi \in \mathit{bad}[n]$
        \State $O \leftarrow \text{SolveNode}(n,\;\mathcal{C}_{\text{node}}(n),\;\pi)$
        \Statex \hspace{4.5em}\textit{// fix edge-relevant props to $\pi$}
        \State $D_{\text{edge}} \leftarrow D_{\text{edge}} \cup O$
        \State $\mathit{valid}[n] \mathrel{+}= \{\pi\}$
        \State $\mathit{bad}[n] \mathrel{-}= \{\pi\}$
        \For{each other node $n'$ in $c$}
            \State $\mathit{bad}[n'] \leftarrow
                   \{\pi' \in \mathit{bad}[n'] \mid
                   \text{UniversallyBad}(\pi',\, n',\, c)\}$
            \Statex \hspace{5em}\textit{// $\pi$ may rescue profiles in $\mathit{bad}[n']$}
        \EndFor
    \EndWhile
\EndFor
\State \Return $D_{\text{edge}}$
\end{algorithmic}
\end{algorithm}